\documentclass[utf8]{FrontiersinHarvard}

\usepackage{url,hyperref,lineno,microtype,subcaption}
\usepackage[onehalfspacing]{setspace}
\usepackage{booktabs}
\usepackage{soul}
\usepackage{xcolor}
\sethlcolor{yellow}
\usepackage{xcolor}

\def\keyFont{\fontsize{8}{11}\helveticabold }
\def\firstAuthorLast{Okkath {et~al.}} 
\def\Authors{Sikha O K\,$^{1,*}$, Abdollah Moslemi\,$^{1,*}$, Elham Alizadeh\,$^{1}$, Ludovic Humbert\,$^{2}$, Miguel A. Gonz\'{a}lez Ballester\,$^{1,3}$ and J\'{e}r\^{o}me Noailly\,$^{1}$}

\def\Address{$^{1}$BCN MedTech, Department of Engineering, Pompeu Fabra University, Barcelona, Spain \\
$^{2}$3D-Shaper Medical
Barcelona, Spain\\
$^{3}$ ICREA, Barcelona, Spain
} 

\def\corrEmail{sikha.okkath@upf.edu, abdollah.moslemi@upf.edu}
\def\corrAuthor{Sikha O K, Abdollah Moslemi}
\def\corrEmail{sikha.okkath@upf.edu, abdollah.moslemi@upf.edu}
\usepackage{soul}

\begin{document}
\onecolumn
\firstpage{1}

\title[GNN- Femur Stress prediction]{
Element-Wise Femoral Stress Prediction: A Hybrid Graph Neural Network Surrogate for Finite Element Analysis}

\author[\firstAuthorLast ]{\Authors} 
\address{} 
\correspondance{} 

\extraAuth{}

\maketitle

\begin{abstract}

\section{}
Hip fractures represent a significant clinical challenge because they cause excess disability and mortality, and have a growing incidence in aging populations, as bone fragility increases. Although finite element analysis (FEA) provides biomechanical assessment of femoral strength, its clinical applicability is mainly limited by high computational cost. Conventional machine learning (ML) approaches have been explored as surrogate models for FEA; predictions are fast but fail to capture the spatial and topological structure of finite element meshes. To address these limitations, this study proposes a hybrid Graph Neural Network (GNN)-based framework for femoral stress prediction using graph-structured finite element modeling (FEM) data. The proposed GCN–SAGE–Residual architecture combines graph convolution, neighborhood aggregation, and residual learning to model spatial dependencies and stress propagation across mesh elements. We systematically evaluated 14 models in a primary comparison, spanning classical ML methods and baseline GNN architectures, complemented by a targeted architectural ablation that isolates the contribution of each component of the proposed design. Evaluation results demonstrate that the proposed hybrid architecture achieves the strongest predictive performance among all evaluated models. The proposed hybrid model achieves the best performance with an R² score of 0.8022 and mean absolute error of 2.879 MPa, highlighting the importance of incorporating mesh topology in biomechanical prediction tasks. The proposed framework provides a scalable and computationally efficient alternative for FEM-based orthopedic analyses and fracture risk assessment. The code for this is available at\url{https://github.com/sikha2552/GNN-Metamodelling-of--Femur/tree/main}:

\tiny
 \keyFont{ \section{Keywords:} Hip fracture, Surrogate Modelling, Graph Neural Networks (GNN), Finite Element Analysis (FEA), Femur Stress Prediction, GraphSAGE} 
\end{abstract}

\section{Introduction}

Hip fractures represent a significant public health burden among elderly populations worldwide. They are frequently associated with long-term disability and elevated rates of excess mortality \citep{Cooper2011}. In Europe, the annual cost of fragility fractures is projected to increase from €37.5 billion in 2017 to €47.4 billion by 2030, while direct fracture care costs for women aged 70 years and older are expected to reach €18.4 billion annually by 2040 \citep{Borgstrom2020,Yeh2023}. Despite declining age-standardized incidence rates, the global number of hip fractures is projected to nearly double by 2050, mainly due to population aging \citep{Sing2023}. Women account for approximately 70\% of all hip fractures because of accelerated postmenopausal bone loss \citep{Lorentzon2022}. Hip fractures are also associated with considerable mortality and disability, with 20--24\% one-year mortality, and many survivors experiencing long-term loss of mobility and independence \citep{Sing2023,Lin2023}. Hence, accurate identification of individuals at high risk is essential to support the development of targeted and personalized preventive strategies.

Currently, fracture risk assessment in clinical practice largely relies on areal bone mineral density (aBMD) derived from two-dimensional dual-energy X-ray absorptiometry (DXA) scans of the proximal femur and the lumbar spine \citep{Stone2003}, as well as predictive tools such as FRAX and Garvan \citep{Adler2014,Watts2009,McCloskey2012}. While these approaches are widely adopted, they remain insufficient to accurately capture the risk of fragility fractures. In particular, they do not account for key determinants of bone mechanical behavior, including three-dimensional bone geometry, the spatial distribution of bone density, and the external loading conditions associated with fracture events. Consequently, more comprehensive metrics that reflect both the structural and mechanical behavior of bone are needed to improve clinically useful fracture risk prediction.

In this context, finite element analysis (FEA) has emerged as a well-established computational framework to investigate complex biomechanical systems. It has been widely applied to simulate and analyze a broad range of physiological and engineering phenomena, including cardiovascular mechanics, hemodynamic behavior, implant design, orthodontic biomechanics, and surgical planning. By enabling the integration of geometry, material properties, and boundary conditions, FEA provides detailed insight into the mechanical response of biological structures under various loading scenarios \citep{Eshghi2011,Post2015,McCulloch1992,Singh2012,Alizadeh2023}. For hip fracture risk assessment, subject-specific three-dimensional finite element models derived from quantitative computed tomography (QCT) or 3D-DXA incorporate patient-specific structural information within a physics-based modelling framework and have demonstrated improved fracture risk assessment compared with conventional aBMD metrics across multiple studies \citep{Adams2018,Kopperdahl2014,Keaveny2023,Fleps2022,Alizadeh2025,RuizWills2019,Grassi2025,Grassi2023,Gugler2026}. However, despite these advancements, the clinical adoption of FEA-based approaches remains constrained by practical limitations. In particular, QCT-based implementations are associated with elevated cost and additional radiation exposure, while both QCT- and DXA-derived FE models require substantial computational resources, specialized software environments, and expert-driven workflows, which limits their scalability and routine use in clinical practice.

In parallel, recent developments in machine learning (ML), deep learning (DL), and artificial intelligence (AI) have introduced data-driven alternatives capable of approximating FEA outputs with substantially reduced computational cost. Data-driven ML approaches have also shown strong performance in broader clinical applications, including disease risk prediction and orthopedic imaging tasks such as fracture detection and classification from radiographs \citep{Ambale2017,Kim2018,Lindsey2018,Olczak2017,Krogue2020,Ferizi2019}.
 Notably, these neural-network and GNN-based methods have produced robust predictions in bone and fracture applications using cohorts of comparable or smaller size \cite{fadul2026explainable}, \cite{shioji2017artificial},\cite{ulivieri2021bone}. However, these tasks are fundamentally different from FE-surrogate regression, and conventional ML's limitations there do not directly establish its inadequacy as a surrogate for continuous stress-field prediction.

Building on these advances, recent efforts have focused on integrating ML and DL techniques with finite element modeling (FEM) to develop efficient surrogate frameworks for biomechanical analysis. Liang et al. \citep{Liang2018} introduced a DL-based approach able to predict full-field stress distributions by learning nonlinear relationships between patient-specific geometries and FEA-derived outputs, achieving near-simulation accuracy while substantially reducing computational cost. Similarly, Madani et al. \citep{Madani2019} generated a large-scale database of parametric FE simulations to train deep neural networks for predicting maximum von Mises stress in arterial walls, demonstrating that DL models can effectively approximate complex biomechanical mappings. In the context of hip fracture assessment, \citep{Awal2025} proposed a ML-based surrogate framework integrating QCT-based FEA to predict fracture risk index and strain distributions within the proximal femur, showing that ML-driven models can assess fall-induced hip fracture probability and probable fracture locations. Despite these advances, most conventional ML methods rely on handcrafted descriptors or tabular representations extracted from medical images or finite element models. Such representations do not explicitly preserve the spatial organization and topological connectivity of FE meshes, preventing the models from learning the mechanical interactions between neighboring regions. As a result, their ability to accurately predict complex biomechanical responses, such as full-field stress distributions, remains limited. In FEM, mechanical responses are inherently defined over discretized meshes with strong topological dependencies between neighboring elements. Graph neural networks (GNNs) provide a natural framework to address this limitation by operating directly on graph-structured mesh representations and explicitly modelling interactions between connected elements. By representing FE discretization as graphs, where nodes encode geometric and material information and edges capture topological connectivity, these models enable the propagation of information across neighboring elements through message-passing operations while maintaining the underlying mesh connectivity of the system \citep{Pfaff2020}. Extending this paradigm, Wu et al. \citep{Wu2024} developed a GNN surrogate for stress prediction on unstructured meshes, while Fu et al. \citep{Fu2023} incorporated boundary-oriented graph embeddings to jointly capture local neighborhood relationships and global boundary conditions in large-scale systems. Feng and Zhou \citep{Feng2024} proposed a Graph Transformer approach to enhance neighborhood information aggregation and capture more complex interactions, and Wurth et al. \citep{Wurth2024} integrated governing physical equations directly into graph-based training, improving generalization across complex geometries and loading conditions while reducing dependence on large datasets. Despite this progress, none of these studies address patient-specific biomechanical surrogate modelling: they are evaluated on synthetic or engineering meshes rather than clinical anatomical data, and none predict the complete stress tensor for a biological structure such as the proximal femur. Within the specific context of hip fracture risk, the surrogate frameworks described above \citep{Liang2018,Madani2019,Awal2025} rely on conventional neural networks or tabular feature representations rather than graph-structured mesh representations, and typically predict a single derived mechanical quantity (e.g., von Mises stress or a fracture risk index) rather than the complete stress tensor. To our knowledge, no prior work has applied GNN-based surrogate modelling to patient-specific 3D-DXA-derived femoral finite element models, nor predicted the complete element-wise stress tensor in this setting, the specific gap this study addresses.

These limitations motivate the development of graph-based surrogate models that preserve the spatial structure of finite element meshes while enabling computationally efficient stress prediction. To address this gap, this study proposes a hybrid GNN framework for predicting femoral stress directly from patient-specific finite element mesh representations. The proposed architecture combines graph convolutional learning, inductive neighborhood aggregation, and residual feature propagation to capture local neighborhood interactions while efficiently propagating information across complex finite element meshes. By operating directly on graph-structured FE meshes, the proposed framework preserves mesh topology and neighborhood interactions, using element-wise physical and anatomical descriptors rather than mesh-level statistical shape descriptors or hand-engineered summary features. The proposed framework is evaluated against multiple classical machine learning methods and representative GNN architectures for element-wise stress field prediction to investigate its potential as an efficient surrogate for patient-specific 3D-DXA-based finite element analysis.

\section{Materials and Methods}

\subsection{Study Cohort and DXA data}

Hip DXA scans were performed at CETIR Grup Mèdic (Barcelona, Spain) using a Prodigy scanner (GE Healthcare, Madison, WI, USA), following the manufacturer's recommended acquisition protocol. Fracture cases were confirmed through clinical records or follow-up. At the same time, controls were verified through clinical history review and serial DXA scans. Overall exclusion criteria at baseline included prior osteoporotic fracture and any condition other than osteoporosis known to affect bone metabolism \citep{Humbert2020}. A retrospective case-control cohort of 128 Caucasian postmenopausal women was used, comprising 64 subjects with an incident hip fragility fracture and 64 age-matched controls without fracture, with baseline DXA visits recorded between 2000 and 2011. Full demographic and densitometric characteristics of both groups are summarized in Table~\ref{tab:demographics}.

\begin{table}[htbp]
\centering
\caption{Demographic and baseline clinical characteristics of the fracture and control groups. p-values were obtained from Mann--Whitney U tests. aBMD: areal bone mineral density (g/cm$^2$).}
\label{tab:demographics}
\begin{tabular}{lcccc}
\hline
& Age & Height (cm) & Weight (kg) &  aBMD (g/cm$^2$) \\
\hline
Control (n = 64) & $68.9 \pm 9.03$ & $153.2 \pm 6.4$ & $62.57 \pm 7.88$ & $0.849 \pm 0.116$ \\
Fracture (n = 64) & $68.9 \pm 8.92$ & $153.3 \pm 7.3$ & $63.18 \pm 11.16$ & $0.758 \pm 0.105$ \\
$p$-value & 0.957 & 0.892 & 0.722 & $<$0.001 \\
\hline
\end{tabular}
\end{table}

\subsection{2D-to-3D reconstruction}
3D-Shaper software (v2.14, 3D-Shaper Medical, Barcelona, Spain) was used to reconstruct subject-specific 3D-DXA models of the proximal femur from standard hip DXA scans \citep{Humbert2017}. Briefly, the method is based on a Statistical Shape and Appearance Model (SSAM) derived from a quantitative computed tomography (QCT) database, in which femoral shape and volumetric bone density variability are characterized using principal component analysis. The SSAM is iteratively aligned with the DXA scan through a 3D--2D registration process that optimizes shape, density, and pose parameters to minimize the difference between the acquired DXA image and the model-derived projection. This approach reconstructs both the cortical and trabecular bone compartments.
The accuracy of the 3D-Shaper-derived measurements has been previously validated against QCT \citep{Humbert2017,Dudle2023,Bracco2026}.

\subsection{Finite Element Analysis}

\subsubsection{Mesh Generation}

\begin{figure*}[h!]
\centering
\includegraphics[width=\textwidth]{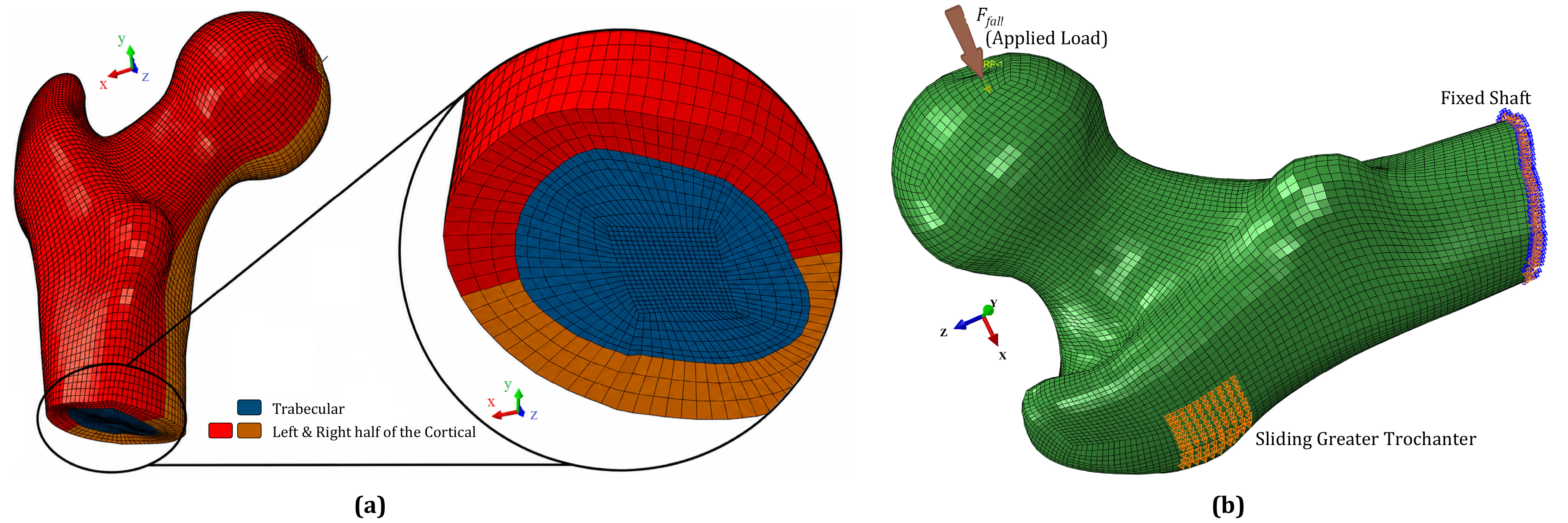}
\caption{The FEM model details: (a) Mesh structure and (b) Load and Boundary Conditions.}\label{fig:00}
\end{figure*}

Patient-specific finite element (FE) models of the proximal femur were generated following a previously established pipeline \citep{Alizadeh2025,RuizWills2019}. A generic hexahedral mesh was first constructed over the mean femur shape derived from the 3D-Shaper\textregistered\ statistical model, and subsequently morphed into each subject's individual geometry using thin-plate splines (TPS). This approach interpolates smooth deformation fields between scattered control points, with 1,110 control points (roughly 10\% of total mesh nodes) sampled from the statistical shape model and matched to the patient-specific geometry provided by 3D-Shaper. The result is a regularized, continuous deformation field applied uniformly across the entire mesh, ensuring one-to-one node correspondence across all 128 cases. Mesh density was selected based on a convergence study comparing three candidate resolutions: 26,200, 126,800, and 334,800 elements. The intermediate resolution of 126,800 hexahedral elements per case produced results comparable to the finest mesh while reducing computation time from 720 to approximately 51 minutes per simulation, making it a practical choice for a dataset of this scale. Within each model, a dedicated layer of three elements was used to represent the cortical shell, preserving the structural distinction between cortical and trabecular compartments (Figure~\ref{fig:00}a). Overall, the mesh structure allowed consistent connectivity, with consistent node and element numbers and IDs over the entire cohort of patient-specific models. Material properties were assigned element-wise using empirical density-elasticity relationships. For cortical bone, the Young's modulus followed $E_{cort} = 10,200 \times {\rho_{ash}}^{2.01}$ MPa, while trabecular bone was characterized by $E_{trab} = 0.003715 \times {\rho_{app}}^{1.96}$. All tissue types were modelled as linearly isotropic elastic with a Poisson's ratio of 0.3, as used and detailed in \citep{Alizadeh2025,RuizWills2019}.

\subsubsection{Boundary Conditions and Loading Scenarios}

The mechanical scenario simulated across all 128 models corresponds to a sideways fall onto the greater trochanter, according to a well-accepted model for the assessment of osteoporotic hip fracture risk, used in the literature \citep{keyak2001effect,RuizWills2019}. Boundary conditions were applied consistently across all cases: the distal end of the shaft was fully constrained, the surface of the greater trochanter was fixed in the direction of the applied force, and a static medio-lateral fall load was distributed across the medial nodes of the femoral head, as illustrated in Figure~\ref{fig:00}b. The fall force was defined in a patient-specific manner, computed from each subject's body mass and height using the expression $F_{fall} = \sqrt{2gh_{c}k_{tissue}m}$, where $g$ is gravitational acceleration ($9.81$ m/s²), $k_{tissue}$ is the unidirectional stiffness of soft tissue overlying the trochanter ($71$ N/mm), $m$ is patient mass, and $h_{c} = 0.51h$ represents the height of the center of gravity \citep{RuizWills2019}. All finite element calculations were performed using ABAQUS, and the resulting element-wise mechanical fields were evaluated separately by anatomical region (femoral neck and trochanteric zone) and tissue type (cortical and trabecular), consistent with the analytical framework that demonstrated the strongest fracture discrimination in prior works \citep{RuizWills2019}.


All finite element simulations were executed in Abaqus 6.12-1, using 8-node linear brick elements (C3D8). Each analysis was configured as a static, general step with geometric nonlinearity enabled (NLGEOM = ON) to account for potential large-displacement effects under the simulated sideways fall loading, and default Abaqus convergence tolerances were applied throughout. The patient-specific fall force was applied as a concentrated load to a single reference node coupled to a rigid-body constraint (Abaqus Interaction Module, Rigid Body) encompassing 316 nodes at the medial surface of the femoral head, thereby distributing the load collectively across that region while preserving its resultant magnitude. The greater trochanter surface was constrained against displacement along the medio-lateral axis only, replicating the sliding contact with the ground surface during a sideways fall, while the distal shaft was fully fixed in all translational degrees of freedom. Only the field outputs of the final increment were requested and recorded. The complete set of six independent Cauchy stress components ($\sigma_{11}$, $\sigma_{22}$, $\sigma_{33}$, $\sigma_{12}$, $\sigma_{13}$, and $\sigma_{23}$) was selected as the primary output for surrogate model training, as it preserves full directional mechanical information and allows subsequent computation of derived quantities, such as von Mises stress or principal stresses, without repeating the simulation. Post-processing and stress extraction were performed using a custom Python script interfacing with the Abaqus output database (ODB) via Abaqus 2024, yielding one stress tensor per element at the centroid level. This element-centroid representation maps directly onto the graph node targets described in Section 2.5, ensuring a strict one-to-one correspondence between simulation outputs and GNN prediction targets with no interpolation. Readers are referred to \citep{RuizWills2019} for further details on the overall finite element modeling pipeline on which the current implementation is based.

\subsection{Data Preparation and Feature Extraction}
\label{sec:feature_extraction}
\subsubsection{Data Splitting Strategy}
All data partitioning was performed at the patient level, before feature extraction or graph construction, applied identically to the classical machine learning baselines (Section~2.7.1), the single-operator GNN baselines (Section~2.7.2), and the proposed hybrid model (Section~2.6), so that no patient contributed elements to more than one partition. This yielded 89 training, 19 validation, and 20 test graphs (69.5\%, 14.8\%, and 15.6\%), corresponding to 11,285,200, 2,409,200, and 2,536,000 elements, respectively, confirming a strictly patient-level split with no element-level mixing. Splitting used random patient-level assignment without considering the fracture status (random seed 42), yielding a fracture/control composition of 44/45 in training, 8/11 in validation, and 12/8 in test, consistent with the overall cohort balance of 64 fracture and 64 control subjects. Splitting by element instead of by patient would let data from the same patient leak into both training and test sets, making results look better than they really are. Splitting by patient avoids this, so test results reflect true generalization to new patients. All normalization statistics were computed from the training set alone and applied unchanged to validation and test data.
\subsubsection{Element-Wise Features}
Each element was assigned a feature vector of nine descriptors, which describe the element's local context within the FE model as follows: (1-3) The Cartesian centroid coordinates ($x,y,z$), encoding the element’s 3D position within the femur; (4) apparent density; (5) elastic (Young’s) modulus, derived from density by tissue-specific empirical equations as described in section 2.2; (6) a binary tissue label distinguishing cortical and trabecular bone; (7) an anatomical region label (head, neck, trochanter, or shaft) providing the model with explicit structural context across the four biomechanically different regions of the proximal femur; (8) a boundary condition (BC) label; (9) a mechanical loading label.

Since each hexahedral element shares its 8 nodes with neighboring elements, different elements touch different numbers of constrained or loaded nodes. A simple binary encoding would lose this detail. Instead, we use a graded scale: the boundary-condition label ranges from 0 (no constrained nodes) to 0.6 (four constrained nodes), and the loading label ranges from 0 to 0.95 based on how many nodes are subjected to the applied load. This gives the model more precise spatial information about how boundary effects vary near the loaded and constrained regions.
All 128 patient models share the same mesh structure (Section 2.2), i.e., element \textit{n} is always in the same anatomical location across all the  patients. We use this shared structure to standardize the graphs: every patient gets the same anatomical labelling, the same graph connectivity, and the same feature definitions. This way, any differences the model sees between patients reflect real biomechanical variation, not inconsistencies in how the graphs were built.

\subsection{Graph Construction}
\label{sec:Graph Construction}

To fully leverage the spatial structure of the femur finite element (FE) mesh,
each mesh was represented as a graph $G = (V, E)$, where each node $v \in V$
corresponds to an individual FE model element and each edge $e \in E$ encodes
direct connectivity among elements. Two elements sharing a common face were
connected by an undirected edge. This representation preserves the anatomical
topology and local interactions that govern stress propagation. Each graph node carried the
nine-descriptor feature vector defined in Section.2.4.2, with the six Cauchy stress components
($\sigma_{11}, \sigma_{22}, \sigma_{33}, \sigma_{12}, \sigma_{13}, \sigma_{23}$)
as prediction targets, computed via FEA.
 
Graph statistics across the 128 femur meshes revealed highly consistent
structural properties (Table~\ref{tab:Graph Statistics}). Each graph contained
exactly 126{,}800 nodes (elements) and 375{,}240 edges, reflecting uniform mesh
discretization across all cases. The average node degree was $5.92 \pm 0.28$
(range 4--6), consistent with the hexahedral mesh topology, in which interior
elements share faces with six neighbors while boundary elements have fewer. The
degree distribution exhibited very low variance (CV = 0.047), indicating high
mesh-quality uniformity. Graph density was $4.67 \times 10^{-5}$ (sparsity
99.995\%), reflecting the sparse, locally connected nature of FE meshes and
enabling efficient sparse matrix operations. All graphs formed a single
connected component, ensuring global information propagation during GNN training. The low average clustering coefficient (0.001) reflects the locally grid-like adjacency structure typical of volumetric hexahedral meshes, in which two face-adjacent neighbors of a given element rarely share a face with each other, rather than indicating a hierarchical or tree-like topology. Degree assortativity was high ($0.787$), indicating that elements tend to connect to
others of similar degree (interior-to-interior, boundary-to-boundary). Receptive-field analysis showed that each element aggregates information from
approximately 6 neighbors at 1-hop, 18 at 2-hop, and 42 at 3-hop. Across the
evaluated architectures, which use 3--6 message-passing layers, the effective
receptive field extends to hundreds of neighboring elements, sufficient to capture stress-propagation patterns across local anatomical subregions while
remaining computationally manageable. The edges were encoded in a sparse adjacency matrix
$A \in \mathbb{R}^{N \times N}$, where $A_{ij} = 1$ if elements $i$ and $j$ share
a face and $0$ otherwise. Owing to the high sparsity (99.995\%), $A$ was stored
in Compressed Sparse Row (CSR) format for memory efficiency. During training,
each node updates its representation by aggregating information from its
neighbors according to $A$, following a message-passing scheme. Formally, the
updated feature $h_i^{(l+1)}$ of node $i$ at layer $l+1$ is
\begin{equation}
\label{eq:message_passing}
h_i^{(l+1)} = \mathrm{UPDATE}\!\left(
  h_i^{(l)},\ \mathrm{AGGREGATE}\left\{
    h_j^{(l)} \mid j \in \mathcal{N}(i)
  \right\}
\right),
\end{equation}
where $\mathcal{N}(i)$ denotes the neighbors of $i$ and
UPDATE/AGGREGATE are layer-specific functions (e.g., GCN, GraphSAGE, or residual
aggregation). This process lets each element integrate information from its local neighborhood iteratively, capturing both local and non-local biomechanical interactions across the mesh.

\begin{table*}[]
\caption{Graph Statistics Across 128 Femur Finite Element Meshes}
\label{tab:Graph Statistics}
\begin{center}
\begin{tabular}{lll}
\toprule
\textbf{Property} & \textbf{Value} & \textbf{Interpretation} \\
\midrule
Nodes per graph & 126,800 & Uniform mesh discretization \\
Edges per graph & 375,240 & Face-adjacency connectivity \\
Average degree & $5.92 \pm 0.28$ & Hexahedral face-adjacency connectivity \\
Degree range & 4--6 & Boundary (4) to interior (6) \\
Graph density & $4.67 \times 10^{-5}$ & Sparse local connectivity \\
Sparsity & $99.995\%$ & Efficient sparse operations \\
Connected components & 1 & Fully connected mesh \\
Clustering coefficient & 0.001 & Locally grid-like connectivity \\
Assortativity & 0.787 & Boundary/interior segregation \\
Edge/Node ratio & 2.96 & Consistent with hexahedral face-adjacency connectivity \\
Receptive field (1-hop) & $5.92 \pm 0.03$ & Immediate neighbors \\
Receptive field (2-hop) & $17.52 \pm 0.13$ & Second-order neighbors \\
Receptive field (3-hop) & $42.47 \pm 0.37$ & Third-order neighbors \\
\bottomrule
\end{tabular}
\end{center}
\end{table*}

\begin{figure*}[h!]
\begin{center}
\includegraphics[width=\textwidth]{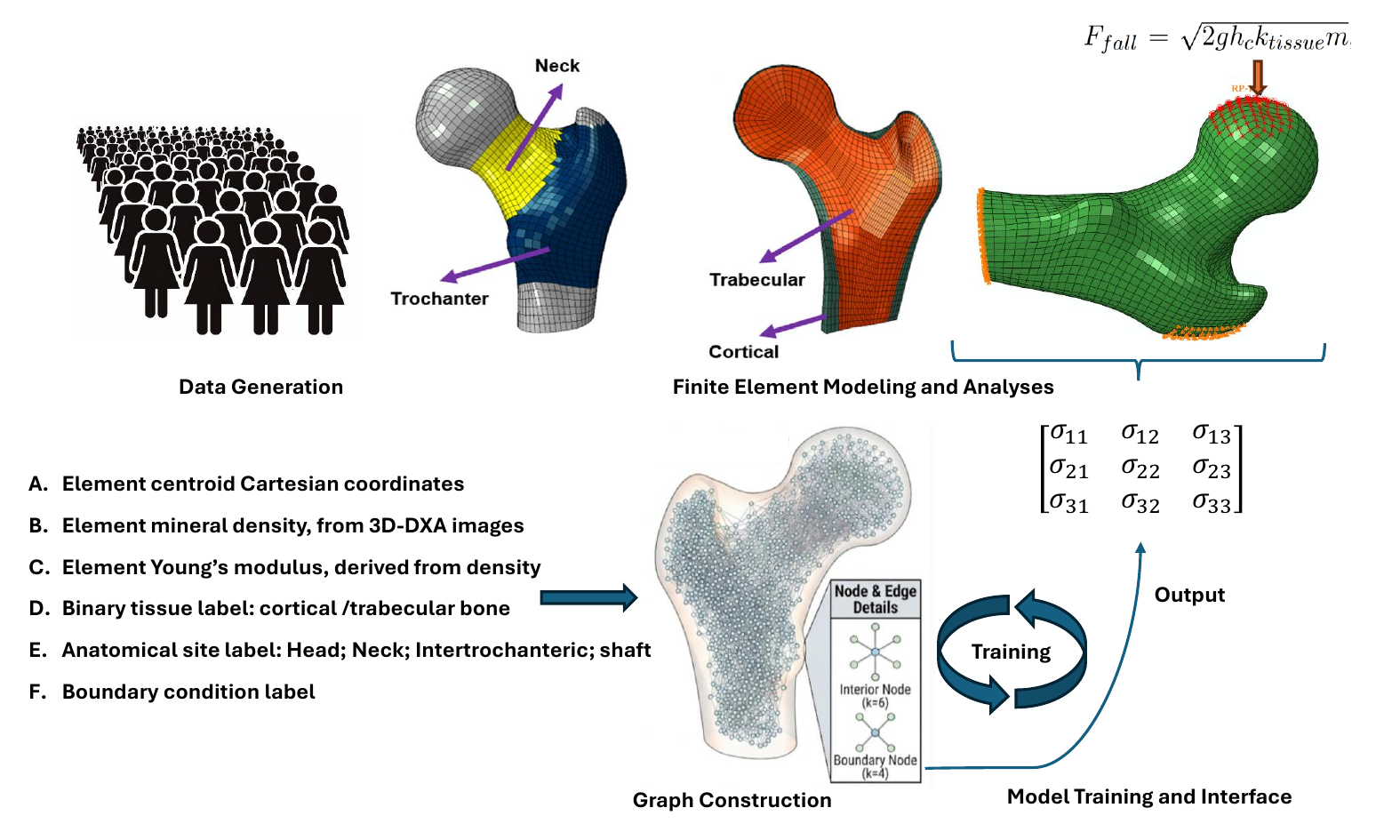}
\end{center}
\caption{ Metamodelling pipeline for element-wise stress prediction in femur}\label{arch}
\end{figure*}
\subsection{Proposed Hybrid GCN–SAGE–Residual Architecture}
\label{sec:proposed}
This section details the proposed hybrid GCN–SAGE–Residual Graph Neural Network surrogate. The model operates on the graph defined in Section~2, where each element is a node and face-sharing elements are joined by edges, and propagates information between connected elements through message passing. This lets each element aggregate information from its neighbors, capturing the spatial stress dependencies that are intrinsic to FE data but inaccessible to models treating elements in isolation. The complete pipeline, FE data generation, feature extraction, graph construction, and surrogate training and inference are illustrated in Figure~\ref{arch}. No single message-passing operator captures the full complexity of stress
propagation in the proximal femur: globally smooth aggregation tends to
over-smooth fine-grained spatial detail, whereas purely local aggregation can
lose structural context across the mesh. To exploit complementary inductive
biases within a single model, we propose a hybrid architecture that combines
graph convolutional aggregation, GraphSAGE aggregation, and residual feature
propagation in every layer.
 
The proposed layer builds on three mechanisms. \emph{Graph convolution} (GCN)
propagates information through symmetric, degree-normalized neighborhood
averaging, which promotes smoothness in the learned representations and captures
gradual stress variations across connected elements. \emph{GraphSAGE} introduces
a learnable aggregation function that explicitly combines each node's own
features with those of its neighbors, providing greater flexibility in
heterogeneous regions where material properties and tissue type vary sharply.
\emph{Residual connections} add a layer's input to its output, so the network
learns incremental feature updates rather than full transformations; this
improves gradient flow, stabilizes training in deeper networks, and preserves
the high-frequency geometric signals that standard graph convolution tends to
wash out. Rather than applying these mechanisms in separate stages, each hybrid layer
fuses them. The updated representation of a node at layer $l+1$ is
\begin{equation}
\label{eq:hybrid}
h^{(l+1)} = h^{(l)} + \sigma\!\left(
    W_1 \cdot \mathrm{GCN}\!\left(h^{(l)}\right)
  + W_2 \cdot \mathrm{SAGE}\!\left(h^{(l)}\right)
\right),
\end{equation}
where $W_1$ and $W_2$ are learnable weights that balance the GCN and GraphSAGE
contributions, $\sigma$ is a nonlinear activation (ReLU), and the additive
$h^{(l)}$ term is the residual connection. The GCN branch supplies globally
normalized neighborhood aggregation that carries structural context across the
mesh; the SAGE branch performs adaptive local aggregation that preserves
element-level feature identity; and the residual term prevents feature
degradation with depth. Together they maintain global smoothness, capture local
heterogeneity, and improve feature fidelity, mitigating the over-smoothing and
vanishing-gradient problems that limit deep graph networks---properties that are
essential for modelling the spatially varying stress fields of patient-specific
femur geometries.
 
The full network stacks six hybrid layers with a hidden width of $256$ channels,
enabling each element to aggregate information over multi-hop neighborhoods (Section \ref{sec:Graph Construction}) while retaining low-level geometric features through the residual paths. Batch normalization is applied after each
graph convolution for stable feature propagation, and dropout ($0.15$) is used
throughout to reduce overfitting. The network predicts all six Cauchy stress
components ($\sigma_{11}, \sigma_{22}, \sigma_{33}, \sigma_{12}, \sigma_{13},
\sigma_{23}$) at every element simultaneously, allowing the model to learn the
internal consistency of the stress tensor across neighboring elements.

\subsubsection{Custom Loss Function}
\begin{figure}[h!]
\begin{center}
\includegraphics[width=10cm]{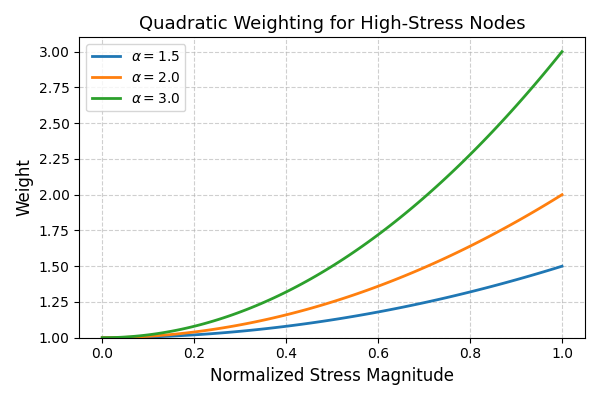}
\end{center}
\caption{ Quadratic MSE Weighting Across High-Stress Nodes for Varying $\alpha$ }\label{loss}
\end{figure}
In fracture stress prediction, errors in high-stress zones are far more critical than errors in low-stress regions. To guide the model to prioritize these regions, we use a weighted Mean Squared Error (MSE) loss that emphasizes high-stress nodes while maintaining stability across the full bone structure. Our \textit{GentleWeightedMSELoss} applies quadratic weighting to target stress magnitudes, offering a less aggressive emphasis than normal MSE loss while still highlighting fracture-prone areas. This works well with the proposed Hybrid GNN (GCN-SAGE-Residual) architecture, which captures both local and global interactions in the bone stress graph.
Let $\mathbf{y}_{\text {pred }} \in \mathbb{R}^{N \times C}$ be the predicted stress vectors and $\mathbf{y}_{\text {true }} \in \mathbb{R}^{N \times C}$ the ground truth, where $N$ is the number of nodes and $C$ is the number of stress components. The standard MSE is:
\begin{equation}
\mathrm{MSE}=\frac{1}{N} \sum_{i=1}^N\left\|\mathbf{y}_{\text {pred }, i}-\mathbf{y}_{\text {true }, i}\right\|^2
\end{equation}
We introduce a weight $w_i$ for each node based on its target stress magnitude:
\begin{equation}
w_i=1+(\alpha-1)\left(\frac{\left\|\mathbf{y}_{\text {true }, i}\right\|}{\max _j\left\|\mathbf{y}_{\text {true }, j}\right\|+\epsilon}\right)^2
\end{equation}
where $\alpha$ is the high-stress weight factor (default $2 \times)$, Squaring the normalized magnitude creates a gentler quadratic curve, avoiding  harsh penalties for high-stress nodes and, $\epsilon$ prevents division by zero. The weighted MSE loss is then:
\begin{equation}
\mathcal{L}_{\text {weighted }}=\frac{1}{N} \sum_{i=1}^N w_i\left\|\mathbf{y}_{\text {pred }, i}-\mathbf{y}_{\text {true }, i}\right\|^2
\end{equation}

This ensures that our hybrid GNN learns both low- and high-stress distributions accurately but pays extra attention to regions likely to fracture. Figure.\ref{loss} illustrates the quadratic weighting applied to high-stress nodes in the GentleWeightedMSELoss for different values of the high-stress factor $\alpha$. The X-axis represents the normalized target stress magnitude, while the Y -axis shows the corresponding weight applied to the MSE. Low-stress nodes remain near a weight of 1 across all curves, ensuring stability in regions with minimal stress, whereas high-stress nodes receive progressively higher weights as $\alpha$ increases. This visualization highlights how choosing a moderate $\alpha$ (e.g., 2.0) gently emphasizes fracture-prone regions without overly penalizing extreme values, balancing sensitivity to critical stress zones with overall training stability.

\subsubsection{Training Strategy}
The proposed large hybrid GNN (GCN–SAGE–Residual) is trained using a carefully designed strategy to ensure numerical stability, efficient computation, and robust generalization for high-dimensional stress prediction.
\begin{enumerate}
    \item \textbf{Optimization and Learning Rate Scheduling}
     
Model parameters are optimized using the AdamW optimizer, which decouples weight decay from gradient updates, improving generalization in deep architectures. The initial learning rate is set to $5 \times 10^{-4}$ with a weight decay of $10^{-6}$. A cosine annealing learning rate scheduler with warm restarts is employed to dynamically adapt the learning rate during training:
\begin{equation}  
\eta_t=\eta_{\min }+\frac{1}{2}\left(\eta_0-\eta_{\min }\right)\left(1+\cos \left(\frac{T_{c u r}}{T_i} \pi\right)\right)
\end{equation}
where $T_0=30$ and $T_{\text {mult }}=2$. This strategy periodically resets the learning rate, allowing the model to escape local minima and improve generalization.
\item \textbf{Mixed-Precision Training}

To improve computational efficiency and reduce GPU memory usage, mixed-precision training is implemented using automatic casting (autocast) and gradient scaling (GradScaler). This enables faster training while maintaining numerical stability, which is particularly important for large GNNs with high-dimensional node features

\item \textbf{Gradient Stabilization and Accumulation}

Training stability is further enhanced through: 1) Gradient clipping with a maximum norm of 1.0 to prevent exploding gradients, 2) Small batch sizes (batch size = 4) due to memory constraints of graph data, 3)Efficient memory management using explicit GPU cache clearing during training. 
While gradient accumulation (typically 8–16 steps) can be used to simulate larger batch sizes, in this implementation we prioritize ultra-stable per-batch updates to avoid instability in deep graph networks.

\item \textbf{Early Stopping and Model Selection}

To prevent overfitting and unnecessary computation, we adopt early stopping based on validation score. Training is terminated if no improvement is observed for a predefined patience of 50 epochs. The best-performing model is selected based on validation performance and saved for final evaluation. This ensures that the model generalizes well to unseen data while avoiding overtraining. Given the depth (6 layers) and capacity (256 hidden channels) of the proposed hybrid GNN, several additional measures are incorporated: Dropout (0.15) applied across layers to reduce overfitting, Batch normalization after each graph convolution layer for stable feature propagation, Periodic memory cleanup to handle large graph batches efficiently.

\item \textbf{Implementation and Reproducibility Details}

All experiments were implemented in Python 3.10.19 using PyTorch 2.0.1 (CUDA 11.8, cuDNN 8700), PyTorch Geometric 2.7.0, scikit-learn 1.7.2, CatBoost 1.2.8, NumPy 1.25.0, and Pandas 2.3.3, on Windows 10. Training was performed on an NVIDIA GeForce RTX 4090 Laptop GPU (17.2 GB VRAM) with a 13th Gen Intel® Core™ i9-13900HX processor (2.20 GHz, 24 cores), 24 physical cores, and 34.2 GB system RAM. All model layers used default PyTorch / PyTorch Geometric initialization, with no custom initialization scheme applied. Training and evaluation hyperparameters for the proposed hybrid model, GNN baselines, and classical machine learning baselines are detailed in Sections~2.6.2, 2.7.1, and 2.7.2, respectively; the data partitioning procedure and random seed are described in Section~2.4.1. 
\end{enumerate}

\subsection{Baseline Models}
\label{sec:baselines}
 
To assess the proposed architecture, we benchmark it against two families of
baselines under an identical feature representation, data split, and training
protocol. Classical machine learning models isolate the value of the graph
representation itself, while single-operator GNNs isolate the contribution of the hybrid design relative to standard graph networks.
 
\subsubsection{Classical Machine Learning Baselines}
\label{sec:baselines_classical}
 
As non-graph baselines, we evaluate classical machine learning models that
operate on the element feature vectors (Section~\ref{sec:feature_extraction})
but treat each element as an independent sample, without access to mesh
connectivity. Because each element is associated with multiple stress outputs,
multi-target regression is handled with the \texttt{MultiOutputRegressor}
framework for models that do not natively support vector-valued predictions. 
Linear models include Ridge, Lasso, and ElasticNet regression. Ridge applies
$L_2$ regularization to stabilize coefficients, Lasso applies $L_1$
regularization to enforce sparsity and implicit feature selection, and
ElasticNet combines both penalties, which is suited to correlated biomechanical
features. We set $\alpha = 1.0$ for Ridge and $\alpha = 0.1$ for Lasso and
ElasticNet, with an $L_1$ ratio of $0.5$ for ElasticNet. To capture nonlinear
relationships, we employ ensemble tree models: a Random Forest regressor
($100$ trees, maximum depth $20$, with minimum-sample constraints) and a
Gradient Boosting regressor ($100$ estimators, learning rate $0.1$, controlled
tree depth), as well as CatBoost. Finally, a Multi-Layer Perceptron (MLP) with
three hidden layers of $256$, $128$, and $64$ units and ReLU activations is
trained with the Adam optimizer (initial learning rate $0.001$, adaptive
scheduling) and early stopping (patience $20$, $10\%$ validation split). All
models use fixed random seeds for reproducibility.
 
\subsubsection{GNN Baselines}
\label{sec:baselines_gnn}
 
To quantify the contribution of the proposed hybrid design, we compare against a
set of single-operator GNN baselines built from the same components introduced
in Section \ref{sec:baselines_classical}. Each baseline shares the feature inputs and
training protocol of the proposed model and predicts the six stress tensor
components at every element using stacked message-passing layers.
 
\begin{itemize}
  \item {GCN (GCN-128, GCN-256):} pure graph-convolutional aggregation
  via normalized neighborhood averaging, evaluated at two hidden widths to
  isolate the effect of capacity. GCNs promote smooth representations but may
  over-smooth fine-grained spatial detail in deeper configurations.
 
  \item {GraphSAGE (SAGE-128, SAGE-256):} learnable aggregation that
  combines self-node and neighbor features, evaluated at two hidden widths.
  Higher-capacity variants better capture nonlinear feature interactions in
  heterogeneous regions.
 
  \item {Graph Attention Network (GAT-128):} attention-weighted
  aggregation that assigns adaptive importance to neighboring nodes, isolating
  the effect of learned attention relative to fixed normalized aggregation.
 
  \item {Residual GCN (GCNResidual-128):} a plain GCN augmented with
  residual connections,
  \begin{equation}
  \label{eq:resgcn}
  h^{(l+1)} = h^{(l)} + \mathrm{GCN}\!\left(h^{(l)}\right),
  \end{equation}
  which isolates the residual mechanism in the absence of the GraphSAGE branch
  of the proposed hybrid layer.
\end{itemize}
 
Comparing these baselines against the proposed architecture
(Equation~\ref{eq:hybrid}) disentangles two distinct sources of performance:
the benefit of operating on graph-structured data, relative to the classical
models, and the benefit of fusing complementary aggregation operators, relative
to the single-operator GNNs.

\subsection{Evaluation Metrics}
To assess the performance of the proposed model in predicting stress distributions, we employ a set of standard regression metrics that capture both accuracy and error characteristics between nodes and stress components.
\begin{enumerate}
    \item {Coefficient of Determination ( $R^2$ )}

The primary evaluation metric is the coefficient of determination $\left(R^2\right)$, which measures how well the predicted values approximate the ground truth. It is defined as:
\begin{equation}
R^2=1-\frac{\sum_{i=1}^N\left(y_i-\hat{y}_i\right)^2}{\sum_{i=1}^N\left(y_i-\bar{y}\right)^2}
\end{equation}
where $y_i$ are the ground truth values, $\hat{y}_i$ are the predictions, and $\bar{y}$ is the mean of the ground truth. An $R^2$ value close to 1 indicates high predictive accuracy. In this work, $R^2$ is used as the primary criterion for model selection and early stopping.

\item {Mean Absolute Error (MAE)}

The Mean Absolute Error (MAE) measures the average magnitude of prediction errors:
\begin{equation}
\mathrm{MAE}=\frac{1}{N} \sum_{i=1}^N\left|y_i-\hat{y}_i\right|
\end{equation}
MAE provides an interpretable measure of average deviation and is less sensitive to outliers compared to squared-error metrics.
\item {Root Mean Squared Error (RMSE)}

The Root Mean Squared Error (RMSE) quantifies the standard deviation of prediction errors:
\begin{equation}
\mathrm{RMSE}=\sqrt{\frac{1}{N} \sum_{i=1}^N\left(y_i-\hat{y}_i\right)^2}
\end{equation}

RMSE penalizes larger errors more strongly, making it suitable for evaluating performance in high-stress regions where large deviations are undesirable.

\item {Per-Component Evaluation}

Since the model predicts multiple stress components (e.g., $\mathrm{S}_{11}, \mathrm{~S}_{22}, \mathrm{~S}_{33}, \mathrm{~S}_{12}, \mathrm{~S}_{13}, \mathrm{~S}_{23}$ ), performance is additionally evaluated per component using $R^2$. This provides a more detailed understanding of how well the model captures different aspects of the stress tensor.
\end{enumerate}
\section{Results}
\subsection{Overall Performance Comparison}
Table.\ref{tab:consolidated} presents a comparative evaluation of classical machine learning (ML) models and Graph Neural Network (GNN)-based architectures for femur stress prediction, using the coefficient of determination (R²) and mean absolute error (MAE) as primary performance metrics.

\subsubsection{Classical Machine Learning Models}
The results indicate that classical machine learning models exhibit limited capability in accurately predicting femoral stress distributions. Linear models, including Ridge, Lasso, and ElasticNet, show very poor performance, with R² values around 0.10–0.11 and high MAE values exceeding 6 MPa. This clearly demonstrates their inability to capture the complex nonlinear relationships between input features and stress responses. Moreover, these models fail to represent spatial dependencies inherent in finite element meshes, resulting in weak predictive power. Ensemble-based models provide noticeable improvements over linear approaches. Gradient Boosting achieves the best performance among classical methods with an R² of 0.7626 and an MAE of 3.246 MPa, followed closely by Random Forest (R² = 0.7617, MAE = 3.177 MPa). CatBoost performs slightly worse (R² = 0.7520, MAE = 3.394 MPa). These models benefit from their ability to capture nonlinear feature interactions; however, they still treat each data point independently and do not incorporate spatial relationships.  The multilayer perceptron (MLP) achieves an R² of 0.7479 with an MAE of 3.371 MPa, performing comparably to ensemble models. While MLPs can model nonlinear mappings, they lack explicit mechanisms to encode mesh connectivity, limiting their effectiveness for FEM-based problems. 
\begin{table*}[h!]
\caption{Comparison of classical machine learning and graph neural network models for femur stress prediction.}
\label{tab:consolidated}
\centering
\begin{tabular}{|l|l|l|l|}
\hline
\multicolumn{1}{|c|}{\textbf{Model}} & \multicolumn{1}{c|}{\textbf{Type}} & \textbf{R²}                            & \textbf{MAE (MPa)}                      \\ \hline
Ridge                                & Linear ML                          & 0.1048                                  & 6.963                                   \\ \hline
Lasso                                & Linear ML                          & 0.1017                                  & 6.888                                   \\ \hline
ElasticNet                           & Linear ML                          & 0.0976                                   & 6.899                                   \\ \hline
Gradient Boosting                    & Ensemble ML                        & 0.7626                                  & 3.246                                   \\ \hline
Random Forest                        & Ensemble ML                        & 0.7617                                  & 3.177                                   \\ \hline
CatBoost                             & Ensemble ML                        & 0.752                                   & 3.394                                   \\ \hline
MLP                                  & Neural (ML)                        & 0.7479                                  & 3.371                                   \\ \hline
GCN-128                              & GNN                                & 0.7300                                  & 3.321                                   \\ \hline
GCNResidual-128                      & GNN                                & 0.737           & 3.302          \\ \hline
GAT-128                              & GNN                                & 0.6835          & 3.645           \\ \hline
SAGE-128                             & GNN                                & 0.7294          & 3.342           \\ \hline
GCN-256                              & GNN                                & 0.7558          & 3.131           \\ \hline
SAGE-256                             & GNN                                & 0.7538          & 3.159           \\ \hline
\textbf{GCN–SAGE–Residual}           & \textbf{Proposed}                  & \textbf{0.8022} & \textbf{2.879} \\ \hline
\end{tabular}
\end{table*}
Overall, even the best classical model (Gradient Boosting) remains  less accurate, highlighting the limitations of non-graph-based learning methods for biomechanical prediction tasks. The fundamental limitation of classical ML models lies in their assumption of independent and identically distributed (i.i.d.) data. In the context of finite element analysis, this assumption is violated because: 1) Stress at a given node depends on its neighboring nodes, governed by physical laws. 2) FEM data is inherently structured as a mesh (graph), not as independent feature vectors. 3) Stress propagation follows spatial correlations and physical constraints, which cannot be captured by tabular models. As a result, classical ML models fail to enforce spatial smoothness and physical consistency, often producing predictions that may be numerically reasonable but physically implausible. Even advanced models like Random Forest and MLP improve performance by capturing nonlinear relationships, but they still lack awareness of topological structure, leading to suboptimal generalization and higher prediction errors.
\subsubsection{Graph Neural Network Models}
Compared to classical machine learning models, some GNN architectures, especially the hybrid architectures, achieve comparable or slightly better performance by leveraging the graph structure of the FEM data. Unlike traditional models that treat each data point independently, GNNs incorporate mesh connectivity, enabling the learning of spatially dependent stress distributions. Baseline GNN architectures, including GCN-128 and SAGE-128, achieve strong predictive performance, with R² values of 0.730 and 0.7294 and low MAE values of approximately 3.321 and 3.342 MPa. By allowing each node to incorporate information from its adjacent nodes, these models effectively capture stress propagation across the femoral structure. The GAT-128 model achieves slightly lower performance (R² = 0.6835, MAE = 3.645 MPa), indicating that attention mechanisms do not provide significant advantages for this application. This can be attributed to the fact that FEM meshes already impose a well-defined spatial structure, reducing the need for learned attention weights.  The GCN-256 model achieves an R² of 0.7558 with an MAE of 3.131 MPa, while SAGE-256 reaches an R² of 0.7538 with an MAE of 3.159 MPa. These results indicate that deeper representations can capture more complex stress patterns; however, simply increasing depth does not fully address the need for efficient feature propagation and stability.

To overcome the limitations of individual GNN architectures, the proposed hybrid GCN–SAGE–Residual model integrates the strengths of multiple graph learning paradigms within a unified framework. This approach achieves the best overall performance, attaining an R² of 0.8022 and the lowest MAE of 2.879 MPa among all evaluated models. The superior performance of the proposed model can be attributed to the complementary roles of its components: 1) GCN layers capture local spatial smoothness, ensuring that stress predictions remain consistent across neighboring nodes. 2) GraphSAGE layers enable robust neighborhood aggregation and inductive generalization, allowing the model to adapt to unseen femur geometries. 3) Residual connections improve gradient flow, stabilize training, and mitigate oversmoothing in deeper architectures. By integrating these mechanisms, the hybrid model effectively learns both local stress continuity and global structural patterns, which are essential for accurate FEM-based predictions. Unlike classical ML models, which ignore spatial dependencies, and standalone GNNs, which capture only partial aspects of the problem, the proposed architecture provides a unified framework that balances expressiveness, stability, and generalization. As a result, the model achieves better results over both classical and baseline GNN approaches, demonstrating the effectiveness of hybrid graph-based learning for biomechanical stress prediction.
\subsection{Architecture Analysis}
\begin{table}[]
\caption{ Performance comparison of GNN architectures for element-wise femur stress prediction. Bold denotes best performance. All metrics are computed on the test set.}
\label{tab:gnnstress}
\begin{tabular}{|l|l|l|l|l|l|}
\hline
\multicolumn{1}{|c|}{\textbf{Model}} & \multicolumn{1}{c|}{\textbf{Architecture Type}} & \multicolumn{1}{c|}{\textbf{\#Parameters}} & \multicolumn{1}{c|}{\textbf{RMSE}} & \multicolumn{1}{c|}{\textbf{MAE}} & \multicolumn{1}{c|}{\textbf{R²}} \\ \hline
GCN-128                              & Single (GCN)                                    & 94406                                      & 8.402      & 3.321     & 0.7300                             \\ \hline

GCNResidual-128                      & Single (GCN+Res)                                & 127942                                     & 8.469                              & 3.302                             & 0.737                            \\ \hline
 
GAT-128                              & Single (GAT)                                    & 95942                                      & 9.217                              & 3.645                             & 0.6835                           \\ \hline

SAGE-128                             & Single (SAGE)                                   & 177862                                     & 8.55                               & 3.342                             & 0.7294                           \\ \hline

GCN-256                              & Single   (GCN)                                  & 369030                                     & 8.089                              & 3.131                             & 0.7558                           \\ \hline

SAGE-256                             & Single (SAGE)                                   & 699782                                     & 8.101                              & 3.159                             & 0.7538                           \\ \hline
 
\textbf{GCN-SAGE-Residual}           & \textbf{Hybrid}                                 & \textbf{709318}                            & \textbf{7.219}                     & \textbf{2.879}                    & \textbf{0.8022}                  \\ \hline
\end{tabular}
\end{table}

\begin{figure}[h!]
\begin{center}
\includegraphics[width=\textwidth]{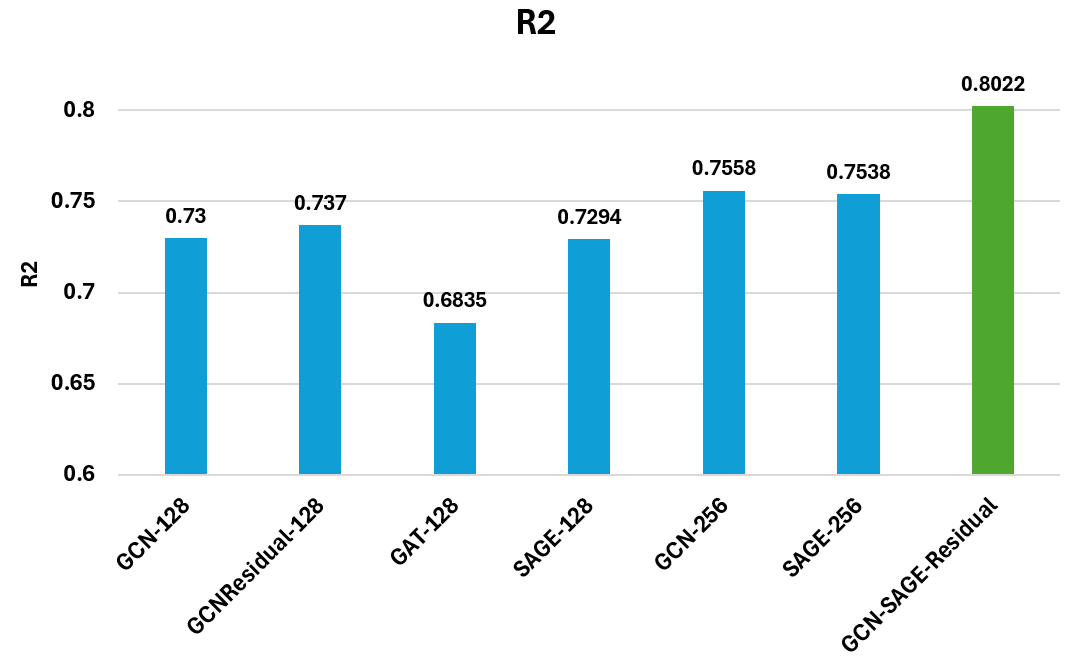}
\end{center}
\caption{ R² scores across GNN architectures for element-wise femur stress prediction. GCN-SAGE-Residual (hybrid) achieves the highest R² = 0.8022, outperforming all single-architecture} variant \label{gnn-stressfig}
\end{figure}
The design of a graph neural network, including its message-passing operator, channel dimensionality, and use of skip connections, fundamentally determines its capacity to capture the spatially complex stress distributions in the femur bone. This section investigates how architectural design choices in graph neural networks influence the accuracy of stress prediction (Table \ref{tab:gnnstress}, Figure \ref{gnn-stressfig}).\\

\textbf{Effect of Hidden Channel Width:} Increasing the hidden dimensionality from 128 to 256 channels produced consistent gains across single-architecture models. GCN improved from R\textsuperscript{2} = 0.730 to R\textsuperscript{2} = 0.7558 (+3.5\%), while GraphSAGE improved from R\textsuperscript{2} = 0.7294 to R\textsuperscript{2} = 0.7538 (+3.3\%) but increased the parameter count by approximately a factor of four (e.g., 177,862 → 699,782 for GraphSAGE; 94,406 → 369,030 for GCN). These results suggest that simply making the model larger is not the primary bottleneck in capturing the complex, spatially varying stress distributions characteristic of patient-specific femur geometries.\\

\textbf{Effect of aggregation strategy:} GraphSAGE and GCN achieved closely comparable performance at their respective 128-channel widths (R\textsuperscript{2} = 0.7294 and 0.730, MAE = 3.342 and 3.321 MPa, respectively). This indicates that mean-pooling aggregation and symmetric normalized convolution offer similar inductive biases for this mesh topology. Adding residual skip connections to a plain GCN (GCNResidual-128, R\textsuperscript{2} = 0.737, MAE = 3.302 MPa) yielded only a marginal improvement over its plain version(GCN-128, R\textsuperscript{2} = 0.730, MAE = 3.321 MPa), indicating that residual connections alone provide limited benefit.\\

\textbf{Attention mechanism under-performance:} Graph Attention Networks (GAT-128) yielded the weakest result across all configurations (R\textsuperscript{2} = 0.6835, MAE = 3.645 MPa). Although attention models can learn different importance for each neighbor, the FEM hexahedral meshes are dense and fairly uniform in structure, so there is not much useful variation for the attention mechanism to learn from. In addition, the multi-head attention design makes the model harder to train and may need more data or stronger regularization to perform well, which was limited in this study.\\

\textbf{Hybrid architecture superiority:} The proposed GCN-SAGE-Residual model achieved the best performance across all metrics (R\textsuperscript{2} = 0.8022, MAE = 2.879 MPa, RMSE = 7.219 MPa), improving on the strongest single-operator baseline (GCN-256, R\textsuperscript{2} = 0.7558) by 6.1\% in R\textsuperscript{2}. This gain reflects the complementary roles of its three components. Because no single message-passing operator captures all three behaviors, combining them in one architecture is a more effective strategy than scaling any one operator alone.\\

\textbf{Parameter efficiency:} Despite its superior accuracy, GCN-SAGE-Residual requires 709,318 parameters, approximately 7.5$\times$ more than GCN-128 (94,406 parameters), which achieves R\textsuperscript{2} = 0.730. This tradeoff merits consideration in deployment contexts: for applications where model size or inference latency is constrained, such as real-time decision support, GCN-128 offers a lightweight alternative that retains competitive predictive fidelity at a substantially reduced computational cost.\\

\textbf{Architectural Ablation:}
To isolate the individual contributions of graph convolution, GraphSAGE aggregation, and residual connections, we evaluated all combinations of these three mechanisms under an identical training protocol (Section~2.7.2), data split, and normalization procedure (Section~2.4.1). This ablation introduces two configurations beyond the fourteen models of the primary comparison (SAGE+Residual-128 and GCN+SAGE without residual connections); the remaining configurations also appear in Tables 3 and 4. Table~\ref{abla} reports parameter counts alongside R\textsuperscript{2}, MAE, and RMSE for each configuration.

\begin{table}[h]
\centering
\caption{Architectural ablation isolating the contribution of GCN, GraphSAGE, and residual connections.}
\label{abla}
\begin{tabular}{lrrrr}
\toprule
\textbf{Configuration} & \textbf{\#Params} & \textbf{R\textsuperscript{2}} & \textbf{MAE (MPa)} & \textbf{RMSE (MPa)} \\
\midrule
GCN-128 (single) & 94,406 & 0.7300 & 3.3210 & 8.4020 \\
SAGE-128 (single) & 177,862 & 0.7294 & 3.3421 & 8.5500 \\
GCN+Residual-128 & 127,942 & 0.7370 & 3.3021 & 8.4690 \\
SAGE+Residual-128 & 179,526 & 0.7140 & 3.5263 & 8.8733 \\
GCN+SAGE, No Residual-256 & 640,198 & 0.7507 & 3.1210 & 8.2238 \\
GCN-SAGE-Residual (proposed) & 709,318 & 0.8022 & 2.879 & 7.2190 \\
\bottomrule
\end{tabular}
\end{table}

Two findings are notable. First, adding residual connections to a single operator does not uniformly improve performance: while GCN+Residual-128 modestly outperforms plain GCN-128 (R\textsuperscript{2} 0.7370 vs. 0.7300), SAGE+Residual-128 performs \textit{worse} than plain SAGE-128 (R\textsuperscript{2} 0.7140 vs. 0.7294), indicating that residual connections alone do not consistently benefit every base operator in this shallow-network setting. 
Second, the GCN+SAGE configuration without residual connections ($R^2 = 0.7507$) performed slightly below both GCN-256 ($R^2 = 0.7558$) and SAGE-256 ($R^2 = 0.7538$), showing that combining the two aggregation operators without residual connections does not by itself exceed the
stronger single-operator baselines. Neither the GCN+SAGE fusion nor residual connections alone accounts for the proposed model's improvement; rather, the three mechanisms are effective in combination, consistent with their complementary roles described in Section~\ref{sec:proposed}. Because these configurations are not parameter-matched, we interpret this as evidence that the combination is beneficial rather than that each mechanism is strictly necessary.

\begin{figure*}[h!]
\begin{center}
\includegraphics[width=\textwidth]{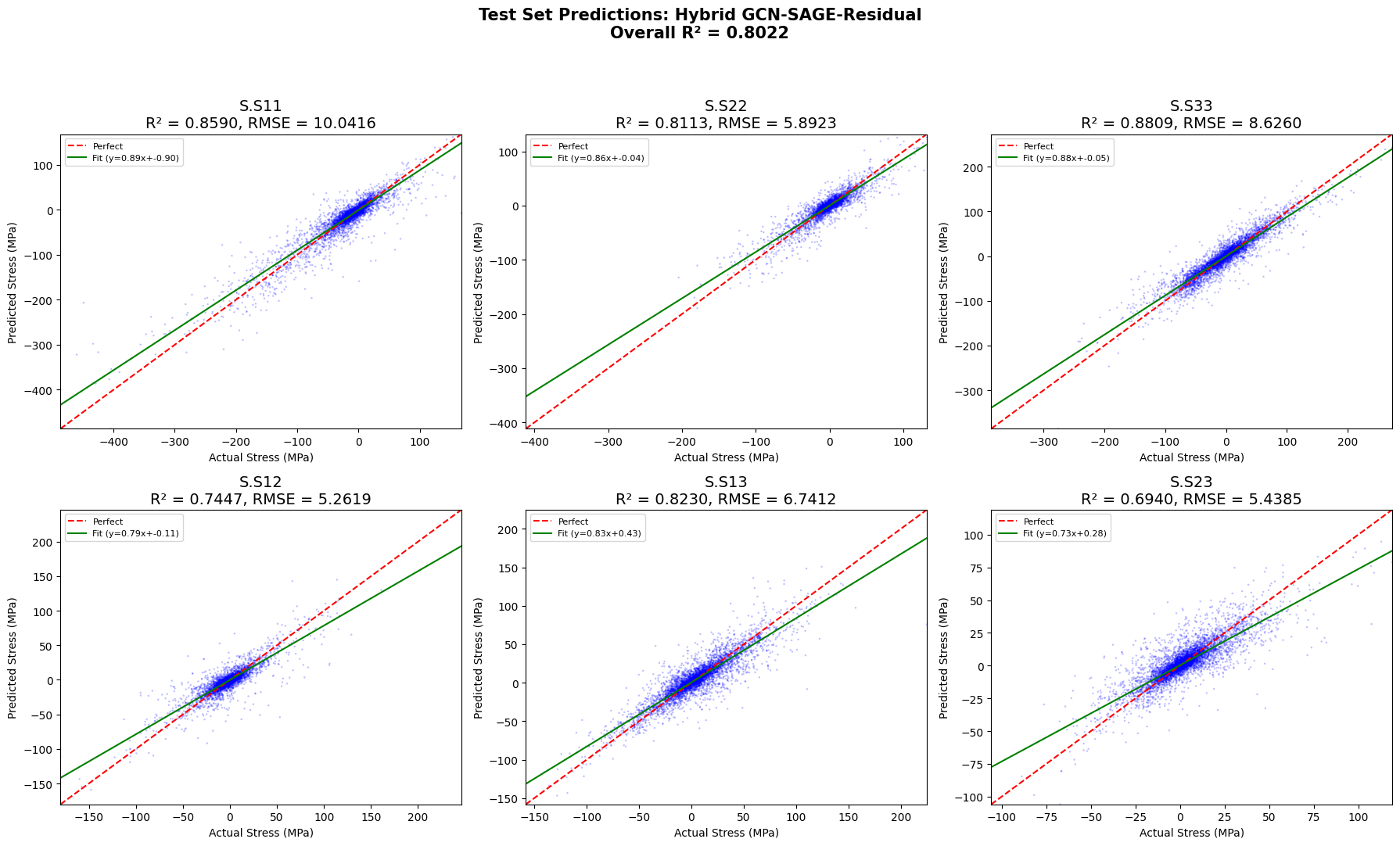}
\end{center}
\caption{ Scatter plot for the per-component stress prediction (R²) for the proposed GCN-SAGE-Residual (hybrid) model }\label{scattter}
\end{figure*}
\subsection{ Per-Component Stress Prediction Results}
To assess the model's capacity to reconstruct the full stress tensor, the predictive accuracy of the GCN-SAGE-Residual model is evaluated independently for each of the six stress components-S11, S22, S33, S12, S13, and S23  on the test set. The per-component R\textsuperscript{2} and RMSE values are reported in Figure~5. The model demonstrates strong predictive performance across the six stress components, with R\textsuperscript{2} values ranging from 0.6940 to 0.8809. The normal stress component S33 achieves the highest accuracy (R\textsuperscript{2} = 0.8809), followed by S11 (R\textsuperscript{2} = 0.8590) and S13 (R\textsuperscript{2} = 0.8230). The remaining components: S22 (R\textsuperscript{2} = 0.8113), S12 (R\textsuperscript{2} = 0.7447), and the out-of-plane shear component S23 (R\textsuperscript{2} = 0.6940) show comparatively lower, though still moderate-to-strong, predictive accuracy. The relatively lower performance on S23 suggests that out-of-plane shear stress is somewhat more difficult for the model to reconstruct than the normal stress components, potentially reflecting its smaller overall magnitude and correspondingly lower signal-to-noise ratio relative to prediction error. Across all six components, the fitted regression slope is consistently below unity (ranging from 0.73 to 0.89), indicating a tendency to under-predict the magnitude of extreme stress values, a pattern consistent with the model's overall regression-to-the-mean behavior noted in Section~3.1.

\subsection{ Qualitative Visualization and High-Stress Region Detection}
\begin{figure*}[h!]
\begin{center}
\includegraphics[width=\textwidth]{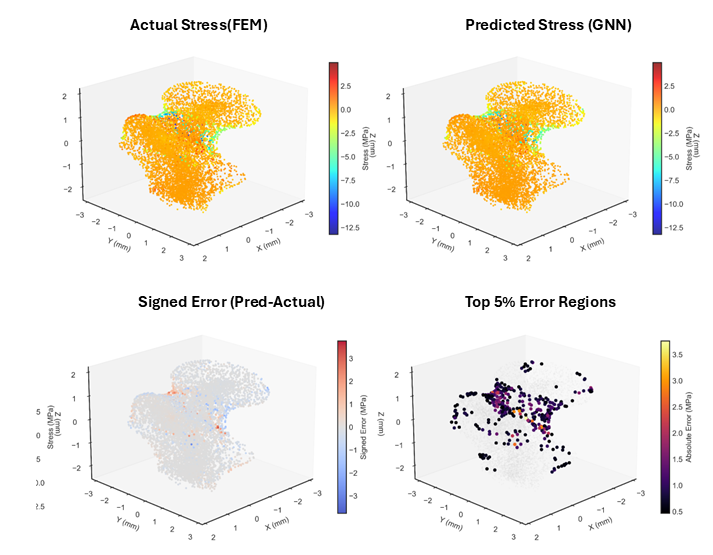}
\end{center}
\caption{ Spatial visualization of predicted versus FEM-derived stress fields, signed prediction error, and top 5\% error regions on a representative test-set femur }\label{stress_pred}
\end{figure*}
To further evaluate the spatial prediction capability of the proposed framework, qualitative visualization experiments were conducted by comparing the predicted stress distributions against the ground-truth finite element method (FEM) results on unseen femur samples. Figure.\ref{stress_pred} presents a representative example of the 3D stress visualization  of the proposed GCN–SAGE–Residual model. The first two panels compare the actual FEM-derived stress field and the stress field predicted by the proposed graph neural network. A strong visual agreement is observed between the two distributions, particularly in capturing the global stress patterns across the femoral geometry. The model successfully reconstructs the spatial variation of stress magnitude, including regions of elevated stress concentration near anatomically critical structures. It is clear from the plot that the overall structural consistency with the FEM ground truth is preserved. The third panel illustrates the signed error distribution (Predicted-Actual), highlighting both overestimation and underestimation regions. The error remains largely centered around zero across most of the geometry, indicating stable and unbiased learning behavior. Localized deviations are primarily observed in geometrically complex regions, such as curved surfaces and boundary-adjacent areas, where stress gradients are inherently steep. To further assess predictive reliability in clinically relevant regions, the final panel highlights the top 5\% of elements ranked by absolute error magnitude. Rather than being uniformly distributed, these largest errors are spatially concentrated in localized anatomical zones—predominantly geometrically complex regions with steep stress gradients, where surrogate approximation is inherently most challenging. 

\subsubsection{High-Stress Region (Hotspot) Detection Analysis}
\begin{figure*}[h!]
\begin{center}
\includegraphics[width=\textwidth]{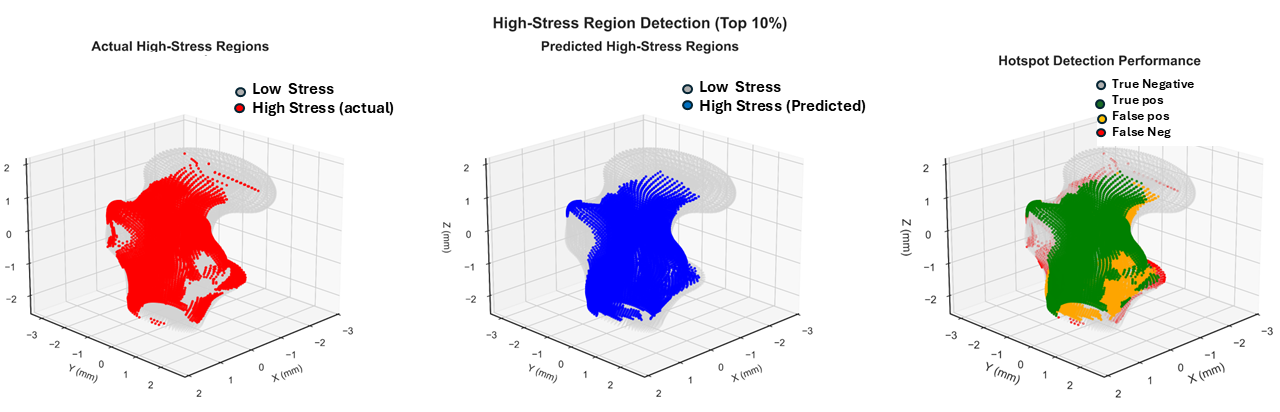}
\end{center}
\caption{ Spatial Evaluation of High-Stress Region Detection Using FEM Ground Truth and GNN Predictions }\label{hotspot}
\end{figure*}
To evaluate the clinical relevance of the proposed GNN-based stress prediction framework, the regression output is further reformulated into a spatial hotspot detection problem. This enables the identification of localized regions of elevated mechanical stress, which may be relevant to regions of elevated mechanical demand in bone structures. The scalar stress measure $\sigma_i$  used for hotspot thresholding is the von Mises equivalent stress, computed from the six predicted (or ground-truth) Cauchy stress components at each element. Let $\sigma_i$ denote the stress value at node $i$. High-stress regions are defined using a percentile-based adaptive thresholding strategy, where nodes belonging to the top $10 \%$ of stress values are classified as hotspots. Formally, a node $i$ is labeled as a hotspot if $\sigma_i>P_{90}(\sigma)$, where $P_{90}(\sigma)$ represents the 90 th percentile of the stress distribution. This adaptive thresholding ensures robustness across different anatomical samples and loading conditions by avoiding the use of fixed global thresholds. The same percentile-based criterion is consistently applied to both finite element method (FEM) groundtruth stress fields and GNN-predicted stress fields, resulting in two binary spatial masks, $H_{\text {FEM }}$ and $H_{\text {GNN }}$, representing true and predicted high-stress regions, respectively. Node-wise comparison between these masks enables the definition of standard classification outcomes: true positives (TP) correspond to $H_{\text {FEM }} \cap H_{\text {GNN }}$, false positives (FP) correspond to $H_{\text {GNN }} \backslash H_{\text {FEM }}$, false negatives (FN) correspond to $H_{\text {FEM }} \backslash H_{\mathrm{GNN}}$, while true negatives (TN) include all remaining nodes. Based on these definitions, standard classification metrics are used to quantify hotspot localization performance. Precision and recall are defined as:
$$
\begin{gathered}
\text { Precision }=\frac{T P}{T P+F P} \\
\text { Recall }=\frac{T P}{T P+F N}
\end{gathered}
$$

The harmonic mean of precision and recall, the F1-score, is computed as:

$$
F 1=\frac{2 \cdot \text { Precision } \cdot \text { Recall }}{\text { Precision }+ \text { Recall }}
$$
These metrics jointly capture both the correctness and completeness of predicted high-stress regions. Quantitatively, the proposed method achieves strong hotspot detection performance, with Precision = Recall = F1-score = 0.8368. Because hotspots are defined as the top 10\% of elements by von Mises stress magnitude for both ground-truth and predicted fields, the number of predicted and actual hotspot elements is identical by construction; precision and recall are therefore mathematically equal under this thresholding scheme, and both pooled (across all test-set elements) and per-patient macro-averaged computations yield the same value (0.8368), indicating consistent hotspot localization performance across individual patients rather than aggregate performance driven by a subset of easy cases. As shown in Figure. \ref{hotspot}, the proposed model effectively captures the dominant spatial distribution of high-stress regions observed in FEM simulations, with a substantial overlap between predicted and ground-truth hotspots indicating strong spatial consistency. Remaining false-positive and false-negative regions highlight localized prediction uncertainties, particularly in structurally complex anatomical areas. Importantly, this comparison provides direct anatomical interpretability of model errors by identifying regions where prediction performance degrades, offering insight into mechanically sensitive zones. 

Hotspot detection evaluates how well the model localizes FE-derived high-stress regions, offering a spatial complement to point-wise error metrics. It reflects agreement with FE-derived stress percentiles, not validated fracture risk or clinical outcomes; establishing that link would require separate validation against fracture criteria or patient outcome data, which is beyond this study's scope. Nonetheless, the high-stress regions identified here, concentrated at the femoral neck and intertrochanteric zone, correspond anatomically to the regions most prone to plastic yielding under von Mises yield surface criteria, which were shown to provide the strongest fracture vs. control discrimination in elasto-plastic FE models of the same cohort \citep{Alizadeh2025}, lending indirect biomechanical support to the spatial relevance of the hotspot analysis.
 
\subsubsection{Fracture vs. Control Subgroup Performance}
\label{sec:subgroup_performance}

To assess whether predictive performance is maintained in the clinically relevant low-bone-density population, test-set performance was evaluated separately for the 12 fracture and 8 control patients comprising the held-out test set. The proposed model achieved R\textsuperscript{2} = 0.7972 and MAE = 3.1825 MPa on the fracture subset, compared with R\textsuperscript{2} = 0.8146 and MAE = 2.4225 MPa on the control subset. Performance is broadly comparable between groups, with a modest reduction in accuracy for fracture-group patients. Given the limited subgroup sample sizes (12 and 8 patients, respectively) inherent to the unstratified test split (Section~2.4.1), these subgroup estimates should be interpreted as indicative rather than statistically definitive; a stratified or cross-validated evaluation across a larger cohort would be needed to establish this difference with confidence.

\section{Discussions}

This study presents a GNN-based surrogate framework for predicting element-wise femoral stress directly from patient-specific 3D-DXA-derived FE meshes. Compared with conventional machine learning methods and several baseline GNN architectures, the proposed hybrid GCN-SAGE-Residual model achieved the highest prediction accuracy (R\textsuperscript{2} = 0.8022, MAE = 2.879 MPa), offering a modest but consistent improvement over the strongest classical machine learning baseline (Gradient Boosting, R\textsuperscript{2} = 0.7626, MAE = 3.246 MPa) and over all single-operator GNN baselines evaluated. These findings support graph-based learning as a viable alternative to computationally expensive FE simulations, while indicating that the advantage of incorporating mesh topology, though real, is more incremental than previously reported.

Prior work has explored deep learning surrogates for biomechanical stress prediction, including full-field stress estimation from patient-specific geometries \citep{Liang2018}, arterial wall mechanics \citep{Madani2019}, and femoral fracture risk indices from QCT-based FE simulations \citep{Awal2025}, but most relied on tabular or non-graph representations. To our knowledge, GNN-based surrogate modelling for patient-specific 3D-DXA-derived femoral FE analysis, particularly for the complete element-wise stress tensor, remains largely unexplored \citep{Wu2024,Fu2023,Feng2024,Wurth2024}. Among the evaluated graph architectures, the hybrid model's advantage over single-operator baselines (e.g., GCN-256, R\textsuperscript{2} = 0.7558) suggests that combining complementary message-passing mechanisms, globally normalized aggregation (GCN), adaptive local aggregation (GraphSAGE), and residual feature propagation, captures aspects of the stress field that no single operator captures alone \citep{Hamilton2017,He2016}. However, the comparatively narrow margin between the hybrid model and strong classical ensemble baselines indicates that, for this element-wise, atlas-standardized formulation, the benefit of graph-structured learning over well-tuned tabular models is real but limited, and should be interpreted accordingly rather than as a categorical advantage of GNNs over classical ML. Predicting the complete six-component Cauchy stress tensor, rather than a single derived scalar, preserves full directional mechanical information and allows downstream computation of principal stresses, von Mises stress, or tissue-specific failure criteria without repeating the FE simulation \citep{keyak2000prediction,Alizadeh2025}. The element-based graph representation, with elements as nodes and face-sharing connectivity as edges, aligns naturally with the FE formulation itself and is consistent with prior mesh-based graph learning strategies \citep{Pfaff2020,Wu2024}. From a clinical perspective, the proposed surrogate could in principle integrate with existing 3D-DXA-based imaging pipelines to reduce reliance on computationally expensive FE simulation, once validated. Once trained, the GNN predicts complete element-wise stress distributions for a single patient in 181.24 ms on consumer-grade GPU hardware (Table.\ref{tab6}), comprising data preprocessing ($78.74 \pm 5.67$ ms), CPU-to-GPU data transfer ($4.05 \pm 1.35$ ms), and model inference ($98.45 \pm 2.06$ ms), with peak GPU memory usage of 2.43 GB. This represents an approximately 16{,}884$\times$ speed-up relative to the $\sim$51-minute FE simulation it replaces (Section~2.3.1). This measures only GNN inference. The full clinical pipeline, including 3D reconstruction, mesh generation, and material-property assignment, has not yet been timed and would be needed before claiming routine clinical use. Hotspot-detection performance (Precision = Recall = F1 = 0.8368) reflects agreement with FE-derived stress percentiles rather than validated fracture outcomes, and should be interpreted as a measure of spatial consistency with FE-derived stress fields rather than clinical fracture-risk validation.
\begin{table}[h]
\centering
\caption{Computational Efficiency: FE Simulation vs. GNN Surrogate Inference}
\label{tab6}
\begin{tabular}{lll}
\toprule
\textbf{Stage} & \textbf{Time (per patient)} & \textbf{Hardware} \\
\midrule
FE simulation (Abaqus) & 51 min & Section~2.3 \\
Data preprocessing & 78.74 $\pm$ 5.67 ms & Intel Core i9-13900HX (24 cores) \\
Device transfer (CPU $\to$ GPU) & 4.05 $\pm$ 1.35 ms & NVIDIA RTX 4090 Laptop GPU \\
GNN model inference & 98.45 $\pm$ 2.06 ms & NVIDIA RTX 4090 Laptop GPU (17.2 GB VRAM) \\
\midrule
\textbf{Total GNN surrogate time} & \textbf{181.24 ms} & \\
\textbf{Speed-up factor} & \textbf{16{,}884$\times$} & \\
\bottomrule
\end{tabular}
\end{table}
Because the fall force is derived from mass and height but is not supplied to the model, we examined whether adding patient-level information improves prediction. Retraining the proposed model with height, weight, and age appended to the element features — under the same split, normalization, and training protocol — did not improve accuracy (R² = 0.7758 vs. 0.8022 for the geometry/density-only model), indicating that this information is largely redundant with the geometric and density features already available: height and weight are the inputs to the fall-force expression, and the resulting forces vary little across this cohort. A consequence is that the surrogate learns the stress response at the cohort’s characteristic load level; it generalizes across subjects under the present sideways-fall configuration, but would require explicit load conditioning and retraining to transfer to markedly different loading magnitudes, directions, or dynamic scenarios. Although the cohort comprises 128 patients, the model is trained and evaluated element-wise: the 20 held-out patients provide approximately 2.5 million unseen test elements under a strict patient-level split with no leakage (Section 2.4.1), so the reported metrics are estimated over a very large number of independent spatial predictions. To assess the robustness of the reported performance given the cohort
size, we computed a patient-level bootstrap confidence interval by resampling the 20 test patients with replacement (2{,}000 iterations). The proposed model's R\textsuperscript{2} was stable across resamples
(R\textsuperscript{2} = 0.80, 95\% CI 0.75--0.85; bootstrap SD 0.026), indicating that the reported performance is not an artifact of the particular train/test split despite the limited cohort size. The small gap between training, validation, and test performance indicates that the model generalizes to unseen patients without overfitting, and performance is maintained across the two clinically distinct subgroups (fracture R² = 0.797, control R² = 0.815; Section 3.4.2). These observations support the robustness of the results despite the cohort size.
Several limitations should be acknowledged. The model was trained on a single, homogeneous cohort (128 postmenopausal Caucasian women, one imaging site, one loading condition representing a sideways fall), and generalization to other demographics, scanners, or loading scenarios has not been evaluated. The cohort size (n = 128, single site) is modest, and although the element-wise evaluation and subgroup consistency support internal robustness, validation on larger, independent, multi-site cohorts remains essential. The applied fall force is not an explicit model input, and augmenting the inputs with patient-level clinical variables (height, weight, age) did not improve accuracy; predictions are therefore tied to the cohort’s characteristic load level and would not transfer to substantially different load magnitudes or directions without retraining. Because all subjects share a fixed atlas-based mesh topology, the model's demonstrated generalization is to unseen patient geometries within this topology, not to unseen graph structures. Total end-to-end pipeline latency, including 3D reconstruction, mesh generation, and graph construction, has not yet been benchmarked against FE simulation time. External validation on independent cohorts, together with integration of predicted stress fields with clinically validated fracture criteria, remains necessary before clinical claims can be made. In summary, this study shows that graph-based learning offers a real, though moderate, improvement over strong classical ML baselines for element-wise femoral stress surrogate modelling, driven primarily by the complementary combination of graph convolution, neighborhood aggregation, and residual learning. Establishing the practical significance of this improvement, and its relevance to fracture risk assessment, will require statistical validation, broader cohorts, and clinical outcome data.

%

\section{Conclusions and Future Works}
In this study, we proposed a hybrid GNN-based framework for predicting femoral stress directly from graph-structured FE meshes derived from patient-specific 3D-DXA models. The proposed framework addresses the high computational cost of conventional FEA while preserving the spatial topology and neighborhood interactions that are essential for accurately modelling biomechanical behavior. Compared with conventional machine learning methods and representative GNN architectures, the proposed GCN-SAGE-Residual model consistently achieved the highest predictive performance, with an $R^2$ accuracy of 0.8022 and a mean absolute error (MAE) of 2.879 MPa for element-wise stress prediction. In addition, the model demonstrated strong agreement with FE-derived high-stress regions, achieving a precision, recall, and F1-score of 0.8368 in hotspot detection. These findings demonstrate that graph-based learning provides an effective and computationally efficient surrogate for patient-specific biomechanical analysis by directly exploiting the topology of FE meshes. Beyond accurate stress prediction, the proposed framework offers a flexible basis for deriving clinically relevant mechanical indicators without repeating computationally expensive FE simulations. Overall, this work highlights the potential of graph neural networks to accelerate patient-specific biomechanical modelling and represents a promising step toward AI-assisted digital twins for personalized hip fracture risk assessment.
Building on these results, future work will focus on strengthening the baseline comparisons to better position graph-based learning among existing biomechanical surrogate models. In particular, we plan to include a global, non-graph baseline that makes use of cross-patient correspondence in the data, as well as dedicated mesh-based surrogate architectures such as MeshGraphNet-style models, to provide stronger points of comparison. In parallel, we are generating enhanced statistical atlases from additional, independent cohorts to evaluate the proposed network formalism across broader populations and imaging settings,against the capacity of the metamodel to show value in clinical risk estimations, as finite element simulations do. Such a clinically-oriented assessment against the FE ground truth technology will provide the external validation needed to establish the value of the approach in the assessment of the risk of bone fragile fracture.

\section*{Ethics statement}
The studies involving humans were approved by the CETIR Grup Medic Scientific Committee. The studies were conducted in accordance
with the local legislation and institutional requirements. The participants provided their written informed consent to participate in this study.
\section*{Data Availability Statement}
The data analyzed in this study is subject to the following
licenses/restrictions: The data are not publicly available due to
privacy or ethical restrictions. Requests to access these datasets
should be directed to Ludovic Humbert, ludovic.humbert@3d-shaper.com. or  Abdollah.Moslemi, abdollah.moslemi@upf.edu.
\section*{Author Contributions}

SO: Conceptualization, Formal Analysis, Software, Methodology, Validation, Writing original draft, AM: Conceptualization, Formal Analysis, Investigation, Methodology, Software, Validation, Visualization, Writing original draft, Writing review and editing. EA: Conceptualization, Investigation, Visualization, Writing – original draft, Writing – review and editing. LH: Supervision, review, and editing. MG: Supervision, review, and editing. JN: Conceptualization, Funding acquisition, Investigation, Methodology, Project administration, Resources, Supervision, Validation, Writing – review and editing.

\section*{Funding}
The author(s) declare that financial support was received for the research and/or publication of this article. This study was funded by the Spanish Government (CPP2021-008574 and CPP2023-010970) and AGAUR (2023 BP 00224).
\section*{Conflict of Interest Statement}

The authors declare that the research was conducted in the absence of any commercial or financial relationships that could be construed as a potential conflict of interest. 

\section*{Generative AI statement}
During the preparation of this work, the author(s) used claude.ai for  language editing and improving the readability of the manuscript. After using this tool, the author(s) reviewed and edited the content as needed and take(s) full responsibility for the content of the publication.
\bibliographystyle{Frontiers-Harvard} %

\bibliography{test}




\end{document}